\documentclass{amsart}
\usepackage{graphicx}
\usepackage{subfigure}

\newtheorem{thm}{Theorem}[section]
\newtheorem{cor}[thm]{Corollary}

\theoremstyle{definition}
\newtheorem{defn}[thm]{Definition}
\theoremstyle{remark}

\numberwithin{equation}{section}

\newcommand{\Real}{\mathbb R}
\renewcommand{\epsilon}{\varepsilon}
\newcommand{\emdp}{$\epsilon$-MDP}

\begin{document}

\title[Plannable RL]{Searching for Plannable Domains can Speed up Reinforcement Learning}
\author{Istv\'an Szita}
\author{B\'alint Tak\'acs}
\author{Andr\'as L\H{o}rincz}

\begin{abstract}
Reinforcement learning (RL) involves sequential decision making in
uncertain environments. The aim of the decision-making agent is to
maximize the benefit of acting in its environment over an extended
period of time. Finding an optimal policy in RL may be very slow.
To speed up learning, one often used solution is the integration
of \emph{planning}, for example, Sutton's \emph{Dyna} algorithm,
or various other methods using \emph{macro-actions}.

Here we suggest to separate \textit{plannable}, i.e., close to
deterministic parts of the world, and focus planning efforts in
this domain. A novel reinforcement learning method called
plannable RL (pRL) is proposed here. pRL builds a simple model,
which is used to search for macro actions. The simplicity of the model
makes planning computationally inexpensive. It is shown that pRL
finds an optimal policy, and that plannable macro actions found by
pRL are \emph{near-optimal}. In turn, it is unnecessary to try
large numbers of macro actions, which enables fast learning. The
utility of pRL is demonstrated by computer simulations.

\end{abstract}
\maketitle

\section{Introduction} \label{s:intro}

\subsection{Reinforcement Learning}

Reinforcement learning involves sequential decision making in
uncertain environments. The sequential aspect of the decision
problem reflects the fact that the immediate cost or benefit of
any state of the environment may play only a small part in
determining the true value of any state. The aim of the
decision-making agent is to develop a \emph{policy}, which
maximizes the benefit of acting in its environment over an
extended period of time.

An often used and efficient framework, which describes
stochastic, sequential decision problems, is the \emph{Markov
decision process} (MDP) (for a review, see
\cite{puterman94markov}). When a problem description satisfies the
requirements of the MDP framework, well-known algorithms can be
used to determine an optimal policy, such as various forms of the
dynamic programming
\cite{bellman57dynamic,bertsekas87dynamic,sutton91planning},
Q-learning \cite{watkins89learning}, or  SARSA
\cite{singh96reinforcement} methods.

These algorithms proceed by maintaining a \emph{value function},
updating it according to the experiences and propagating the
values of states. Under appropriate conditions, RL algorithms are
shown to develop an optimal policy
\cite{bertsekas87dynamic,singh00convergence}. However, the basic
forms of these algorithms propagate the values step-by-step,
therefore convergence may be very slow. Various methods have been
developed to overcome this difficulty, such as prioritized
sweeping \cite{moore93prioritized}, eligibility traces
\cite{Sutton88Learning,singh96reinforcement} and also planning
methods to be described below.

\subsection{Planning}
Besides RL, another successful and well-studied approach to
solving decision problems is \emph{planning}. As a central problem
of classical AI research, effective algorithms have been developed
to solve planning problems. However, classical decision-theoretic methods usually
assume that the world is deterministic. This is an appropriate
approximation in some cases, but \emph{not always}. There may be
certain parts of the world, that are highly stochastic, making
classical planning unreliable, or even useless. It is a plausible
idea to integrate planning and reinforcement learning to gain more
general and robust methods.

Sutton and colleagues as well as others
\cite{sutton91planning,sutton91integrated,peng93efficient,forbes00practical}
have integrated planning and learning in the \emph{Dyna}
architecture: Planning is treated as being virtually identical to
reinforcement learning. Learning updates the appropriate value
function estimates according to experience as it actually occurs,
whereas planning updates the same value function estimates for
simulated transitions chosen from the world model.

Other researchers \cite{sacerdoti77structure,korf85learning} used
planning to develop \emph{macro actions} (i.e., fixed sequences of
actions) that could speed-up value propagation and learning. A
macro action could be a complete sub-policy (such as `go to the
door', `search a wall', etc.) \cite{hauskrecht98hierarchical,
precup98theoretical, mcgovern98macroactions}. The main difficulty
with macro actions is how to construct them: they must be either
handcrafted (see, e.g.,
\cite{kaelbling93hierarchical,Kalmar98Module-based}) or one must
try to generate them automatically (see, e.g.,
\cite{dietterich00hierarchical}). In this latter case a great
number of useless macros might be generated, which might even
deteriorate learning
\cite{mcgovern98macroactions,kalmar99evaluation}.

\subsection{Thesis of the Paper}

We propose a method called plannable RL (pRL), that has
similarities with both planning methods mentioned: we make focused
value updates based on hypothetical experiences gained from a
model. This is similar to the Dyna architecture. On the other
hand, we maintain two separate (but interacting) value functions,
one for learning and one for planning. We use a simple model to
evaluate the value function for planning: values are updated when
transitions are considered plannable (i.e., when those are close
to deterministic). This planning-value function is then used to
compute macro actions. We show that these macro actions are
\emph{near-optimal}. This means that we do not need to generate
large numbers of (possibly bad) macro actions, and fast learning
is still possible.

\subsection{Structure of the Paper}

First, the basis of RL methods are reviewed
(Section~\ref{s:overview}. The pRL algorithm is described and a
pseudo-code of the algorithm is provided in
Section~\ref{s:wp-dyna}. The near optimality of the algorithm is
proven in Section~\ref{s:optimality}. Computational demonstration
are provided in the next section, i.e., in
Section~\ref{s:experiments}. The paper is finished by a discussion
(Section~\ref{s:discussion}).



\section{Preliminaries} \label{s:overview}

\subsection{Reinforcement Learning (RL) and the MDP framework}

Let us recall the definition of a Markov decision process (MDP) \cite{puterman94markov}.
A (finite) MDP is defined by the tuple $\langle X, A, R, P \rangle$, where $X$ and $A$
denotes the finite set of states and actions, respectively. $P: X \times A \times X
\rightarrow [0,1]$ is called the transition function;  $P(x,a,y)$ gives the probability
of arriving at state $y$ after executing action $a$ in state $x$. Finally, $R: X \times
A \rightarrow \Real$ is the reward function, $R(x,a)$ gives the immediate reward for
choosing action $a$ in state $x$.

At each sequence of discrete time steps, $t = 0, 1, \ldots$, the
problem-solving agent observes its world to be in state $s_t \in
S$ and executes an action $a_t \in A$. After this action, the
agent receives reward $r_{t+1} \in \Real$ from the world and
observes the next state, $s_{t+1}$ according to the functions $R$
and $P$.

The agent's objective is to choose each action $a_t$ so as to maximize the expected
discounted return, $\mathcal{R}_t$, i.e., $E\left( \mathcal{R}_t = r_{t+1} + \gamma
r_{t+2} + \gamma^2 r_{t+3} + \ldots \right)$, where $0 \le \gamma < 1$ is the discount
factor and $E(.)$ denotes the expected value of the argument.

A Markovian policy $\pi$ is a $S \times A \to [0,1]$ mapping, where $\pi(x,a)$ is the
probability that in state $x$ the agent selects action $a$. The value of a state $x$
under policy $\pi$ is the expected value of the total discounted reward of starting from
$x$ and then following policy $\pi$. Formally, $V^\pi (x) =
E_{P,\pi}(\sum_{t=0}^{\infty} \gamma^t r_t | s_0 = x)$ (i.e., the expected value depends
on the transition probabilities and the policy applied). These values satisfy the
recursive equations
\[
  V^\pi (x) = \sum_{a\in A} \pi(x,a) \left( R(x,a) + \gamma \sum_{y\in S} P(x,a,y) V^\pi(y) \right) \text{ for all } x.
\]
A policy $\pi^*$ is optimal, if for all policies $\pi$, $V^{\pi^*} (x) \ge V^\pi (x)$ for
all $x \in S$ (it is easy to show that such a policy exists). The
corresponding value function satisfies
\begin{equation} \label{e:V*}
  V^* (x) = \max_{a\in A} \left( R(x,a) + \gamma \sum_{y\in S} P(x,a,y) V^*(y) \right).
\end{equation}

A standard way to find an optimal policy is to compute the optimal
value function $V^*: X \rightarrow \Real$, which gives the value
(i.e., the expected accumulated discounted reward) for each starting state. From $V^*$, the optimal policy can be derived:
the `greedy' policy with respect to the optimal value function is
an optimal policy (see, e.g., \cite{sutton98reinforcement} and
references therein). This is the well-known \emph{value iteration
approach} \cite{bellman57dynamic}. Functions $V^\pi$, $V^*$ are
called value functions, which are associated with a fixed policy $\pi$ and with
the optimal policy, respectively.

We can also define the action value function $Q^\pi(x,a)$ as the
expected value of the total discounted reward of starting from
state $x$, choosing action $a$ and then following policy $\pi$.
This value function is often more useful, because it provides the
values of individual actions. It is easy to see that
\[
  Q^\pi (x,a) = R(x,a) + \gamma \sum_{y\in S} P(x,a,y) V^\pi(y),
\]
\[
  V^\pi (x) = \sum_{a \in A} \pi(x,a) Q^\pi(x,a)
\]
and thus
\[
  Q^\pi (x,a) = R(x,a) + \gamma \sum_{y\in S} P(x,a,y) \sum_{a' \in A} \pi(y,a') Q^\pi(y,a').
\]
The optimal action-value function $Q^*$ satisfies an equation
analogous to $V^*$:
\[
  Q^* (x,a) = R(x,a) + \gamma \sum_{y\in S} P(x,a,y) \max_{a'\in A} Q^*(y,a').
\]

\subsection{Dynamic Programming and Two Basic RL Algorithms}

Equation \ref{e:V*} can be used as an iteration:
\begin{equation} \label{e:V_t}
  V_{t+1} (x) = \max_{a\in A} \left( R(x,a) + \gamma \sum_{y\in S} P(x,a,y) V_t(y) \right)
\end{equation}
for every state $x$, and for an arbitrary $V_0$. The iteration is
known to converge to $V^*$. The method is called \emph{value iteration}.

If the model of the environment (i.e. $R$ and $P$) is not known,
or the state space is too large for solving Eq.~(\ref{e:V_t}),
then sampling methods can be used. One such algorithm is called
\emph{Q-learning} \cite{watkins89learning}. Q-learning uses the
following iteration:
\begin{equation} \label{e:Qlearning}
  Q_{t+1}(s_t,a_t) = (1-\alpha_t) Q_t(s_t,a_t) + \alpha_t \left( r_t + \gamma \max_{a \in A} Q_{t} (s_{t+1},a) \right),
\end{equation}
where $s_t$, $a_t$ and $r_t$ are the state of the system, the
selected action, and the immediate reward at time step $t$,
respectively, $Q_0$ is arbitrary, and $0 \le \alpha_t \le 1$ is
the learning rate.


It has been shown that if $\alpha_t$ converges to 0 properly
($\sum_t \alpha_t$ is divergent, but $\sum_t \alpha_t^2$ is
convergent), and if every $(x,a)$ pair is updated infinitely
often, then $Q_t$ converges to $Q^*$ with probability 1. The proof
can be found, e.g., in \cite{singh00convergence}.

Another RL method is SARSA, which takes sampling to the extreme.
It has an update rule similar to (\ref{e:Qlearning}):
\begin{equation} \label{e:SARSA}
  Q_{t+1}(s_t,a_t) = (1-\alpha_t) Q_t(s_t,a_t) + \alpha_t \left( r_t + \gamma  Q_{t} (s_{t+1},a_{t+1}) \right),
\end{equation}
where $Q$-values of the iteration are action value estimates of
two state-action pairs, the one which just occurred and its
predecessor. SARSA is convergent under the same assumptions as
Q-learning \cite{singh00convergence}. A comprehensive overview of
various reinforcement learning methods can be found in
\cite{sutton98reinforcement} and in the references therein.

\subsection{Planning with Dyna}

Reinforcement learning often requires a large number of experiences (i.e. $(s_t, a_t,
r_t, s_{t+1})$ tuples) to develop an appropriate policy. This is efficient when
experience can be collected quickly, provided that one can afford the cost of
explorations. \emph{Dyna} offers a solution to improve the exploitation of previous
experiences by integrating planning into the learning process.

Informally, planning is a process of computing a (near-)optimal policy for the existing
(possibly inaccurate) model of the environment. Planning is an off-line method, it
improves the policy without invoking additional interactions with the environment. DP,
for example, executed on the available model, is a planning algorithm. Limitation arises
if the model is not given: experience is needed to build a model. Another problem
appears when DP is computationally intensive: performing a single and \emph{complete} DP
iteration requires $O(|S|^2)$ computation steps. Efforts have been made to overcome this
drawback, yielding solutions like \emph{prioritized sweeping} \cite{moore93prioritized},
or \emph{real-time dynamic programming} \cite{Barto95Learning}.

Another successful approach is to combine reinforcement learning
and dynamic programming in a single algorithm. This approach was
first suggested by Sutton, who called the algorithm \emph{Dyna}.
The basic idea is that \textit{one} DP update on \textit{one}
single state can be interpreted as a reinforcement learning step.
In the Dyna architecture the agent repeats the following steps:
(1) obtain experience from the environment, (2) use this to update
the value function, (3) use the experience to improve the model of
the environment (e.g. to approximate the transition
probabilities), (4) obtain \emph{hypothetical experience} from the
updated model, and (5) use the hypothetical experience to update
the value function. Note that Dyna focuses DP iterations on
previously visited (and thus presumably important) regions of the
state space, and therefore it might reduce the computational
demand of DP significantly.

Dyna allows the adjustment of planning relative to collecting
experiences: if there is no time for planning, steps (4) and (5)
can be omitted. Conversely, if real experience is slow or costly,
then multiple planning steps may be accomplished.

\subsection{Planning with Macro Actions}

Another possibility for utilizing additional computational
capacity is to compile compound actions (macro actions) from basic
ones. In its simplest form, macros are fixed sequences of actions
(e.g. `go forward, turn right, go forward'). Many works have dealt
with different versions of the macro concept. In particular, one
might consider policies of sub-problems (e.g., policy for `finding
a wall' or policy for `going to the door') as macros. In this case
separate value functions on separate parts of the state-space will
arise.

In order to integrate macros into reinforcement learning, two
additional steps must be implemented: (6) the generation of macro
actions and (7) the evaluation of macro actions. The latter one
can be accomplished, e.g., by computing $Q(x, a_{\textrm{macro}})$
in all necessary states. Generating appropriate macro actions is a
more difficult problem. One commonly used approach is to pre-wire
them by hand \cite{kaelbling93hierarchical}. This is a
straightforward way to encode prior knowledge into the learning
problem if such prior knowledge is available and if the
environment is steady. Efforts have been made also to generate
macro actions in an automated fashion. The general approach is to
use some heuristics to compile a number of macros from basic
actions, evaluate the macros, and then keep the ones which are the
most useful.

A useful set of macro actions can substantially speed up learning,
because the agent can make larger steps toward its goal. On the
other hand, evaluating bad macros can deteriorate performance
(because trying a bad macro takes a large step in a wrong
direction).

\section{The pRL algorithm} \label{s:wp-dyna}

In this section a novel algorithm is proposed, which combines DP
and macros. First, a model is built for model-based value-function
updates, just like in Dyna. These updates are used directly to
find useful macro actions and to calculate their utility. A
parallel approximate value function is maintained, which encodes
the macros and their values according to the model.
Greedy policy with respect to this approximate value function is
applied. The resulting policy is then used to choose a macro
action.

An important question is the choice of the model. One might
attempt to learn an approximation of the $P(x,a,y)$ transition
probabilities and the corresponding rewards. However, estimating
and maintaining a table of size $|S|^2 \cdot |A|$ may not be
feasible. Finally, we think that planning is more effective in
near-deterministic domains of the state space, and is less
advantageous when little is known about the outcome of an action.
Such problems frequently arise in real life. For example,

\begin{itemize}
  \item when one has to pass a city, one might choose going through the city
  directly or taking the ring around the city. The first option could be less
  expensive in many cases, but large durations on travel time may arise if
  a traffic jam occurs in the inner areas. The second option may
  allow for more accurate planning and may have a competitive cost
  on the average.
  \item Requests can be distributed for mobile agents in many different ways. Planning
  becomes crucial when part of the system may be(come) unreliable.
\end{itemize}

Our planning algorithm (the planner) stores only those transitions
that are almost deterministic, which we shall call \emph{plannable
transitions}.\footnote{Note that we allow for almost deterministic
transitions but we do not assume an almost deterministic
environment here.} The algorithm will be called plannable-RL, or
\emph{pRL}, for short.

\subsection{Basic Definitions, the Value Functions and their Update
Rules}\label{ss:defs}

In the following we specify the pRL algorithm. First of all, we define plannable transitions formally:

\begin{defn}
A transition $(x,y)$ is called \emph{plannable with accuracy
$\kappa$}, if it can be realized with probability $\kappa$, i.e.
there exists an action $a(x,y)$ such that $P(x,a(x,y),y) \ge
\kappa$.
\end{defn}

Let us denote the set of states that are plannable from state $x$
by $T(x)$, i.e.
$$T(x) := \{ y : (x,y) \ \textrm{is $\kappa$-plannable} \} = \{ y \mid \exists a(x,y): P(x,a(x,y),y) \ge \kappa \}$$

We assume that we are given an inverse dynamics $\phi: X \times X
\to A$ such that $\phi(x,y) = a(x,y)$, if $(x,y)$ is plannable,
and it is arbitrary otherwise. This is a reasonable assumption;
either the inverse dynamics is known in advance, or, the `learning
by doing' method can be sufficient to approximate the inverse
dynamics for the plannable domains.

To compute $T(.)$, we approximate $P(x,\phi(x,y),y)$ with
$\hat{P}_t(x,y)$ using the following approximation scheme:
\begin{equation} \label{e:P_hat_update}
  \hat{P}_{t+1}(x,y) =
  \begin{cases}
    (1-\alpha_t) \hat{P}_{t}(x,y) + \alpha_t \cdot 1
        & \text{if $s_t = x$, $a_t = \phi(x,y)$ and $s_{t+1} = y$}, \\
    (1-\alpha_t) \hat{P}_{t}(x,y) + \alpha_t \cdot 0
        & \text{if $s_t = x$, $a_t = \phi(x,y)$ and $s_{t+1} \ne y$}, \\
    \hat{P}_{t}(x,y)
        & \text{otherwise},
  \end{cases}
\end{equation}
$$
  \hat{P}_0 \equiv 1.
$$
Note that this iteration approximates the exact transition
probabilities. A transition $(x,y)$ is considered plannable, iff
$\hat{P}_t(x,y) \ge \kappa$. These almost-sure transitions will be
considered as \emph{sure} by our planning algorithm. The
approximation simplifies the required computations, and -- as we
shall show later -- provides near-optimal solutions.

\begin{defn}
Connected components of plannable transitions are called
\emph{plannable domains}.
\end{defn}

Note that plannable domains are disjoint. In turn, the number of
plannable domains is smaller than the number of states.

The immediate rewards of plannable transitions are similarly
approximated by $\hat{R}(x,y)$.

In our algorithm, two value functions are maintained: a standard
(basic) action-value function $Q(x,a)$, $a \in A$, which is
updated by real experiences, and another value function called the
planning value function. The planning value function will make use
of a Dyna-like algorithm above the simplified model described in
Eq.~(\ref{e:P_hat_update}). Both functions suggest -- possibly
different -- policies. In every time step, pRL can switch between
these policies by examining which one seems better.

\subsection{Updating the basic value function} For updating the basic value function,
any traditional RL update rule can be applied. For example, one may use DP, Q-learning,
SARSA, etc. We use SARSA because of its simplicity:
$$
  Q_{t+1}(s_t,a_t) = (1-\alpha_t) Q_t(s_t,a_t) + \alpha_t \left( r_t + \gamma  Q_{t} (s_{t+1},a_{t+1}) \right),
$$
where $s_t$, $a_t$ and $r_t$ are the state of the system, the
selected action and the reward at time step $t$, respectively.
$Q_0$ is arbitrary, and $0 \le \alpha_t \le 1$ denotes the
learning rate. Note that the basic value of state $x$ at time $t$
is given by $V_t(x) = \max_a Q_t(x,a)$.

\subsection{Updating the planning value function} Several update rules could be used for the planning value function.
However, there is an important difference: an (approximate) model
and an inverse dynamics is known for this case. Therefore
We do not need to maintain an action-value table, instead a simple state value function
  can be used. This function will be denoted by $\hat{V}(x)$. The corresponding policy
  can be determined, e.g., by the inverse dynamics. Note that this
  policy may differ from the policy suggested by the basic value
  function.

We chose the value iteration update rule in the following form:
$$
  \hat{V}_{t+1}(x) = \max_a \sum_y ( P(x,a,y) (R(x,a,y) + \gamma' \hat{V}_t(y) )
  ).
$$
The number of non-vanishing terms of this update may be
considerably smaller than the number of all possible terms,
because in our approximate model all transition probabilities are
either 0 or 1. Transitions with low probability are omitted,
whereas transitions with high probability (determined by $\kappa$)
are considered as sure transitions. Note that this simplification
could mislead action selection. The error depends on the degree of
the simplification, which is determined by $\kappa$: lower
$\kappa$ results in a coarser model. Assuming that immediate
reward $R$ is approximated by $\hat{R}_t$, the update rule can
be rewritten as
\begin{equation}\label{e:hat_V}
    \hat{V}_{t+1}(x) = \max \left\{ \max_{y \in T(x)} (\hat{R}_t(x,y) + \gamma' \hat{V}_t(y) ) , V_t(x) \right\}.
\end{equation}

Here we took into consideration that sometimes it is better to select actions according
to the basic value function $V_t(x)$ (i.e. continue sampling or quit planning and start
sampling) than to select action according to the planning value function $\hat{V}_t(x)$)
(i.e., continue planning, or quit sampling and start planning). Note that when no
plannable states are available, one has to return to the basic value function. One may
think of this technique that the space of plannable actions is extended by a new
pseudo-action, which could be called `\emph{stop planning}': The choice of this action
retreats to sampling and to action selection by means of the basic value function.

The update rule is applied for the planning value action in the
plannable area around the current state $x$ in every time step.
This area can be determined by a limited breadth-first search on
the graph of plannable transitions, starting from state $x$. The
search phase and the update phase require no interactions with the
real system, these are `\textit{off-line}' evaluations. Due the to
limited search and the limited number of DP updates, they take
only $\mathcal{O}(C)$ steps.

\subsection{Action selection} In classical RL methods (Q-learning, SARSA, etc.), $\epsilon$-soft
or $\epsilon$-greedy policies with respect to the actual value
function can serve to support exploration and exploitation (see,
e.g.,  \cite{sutton98reinforcement}). Now we have two different
value functions which may suggest different policies. We shall
make use of the $\epsilon$-greedy policy when $Q_t$ is used, but
if the value of planning from the actual state is higher, i.e.
$\hat{V}_t(x) > V_t(x)$, then action will be generated by means of
the planning value function $\hat{V}_t$. This decision is made by
pRL in every state.

\subsection{Plannable transitions as macros}

Macros are not represented explicitly in pRL, only through the planning value function. This is advantageous because we do not need much space to store the macros - and we can still store the most useful macro with its value for each state.

The macro encoded by $\hat{V}$ is \emph{"choose the greedy action according to $\hat{V}_t$, while $\hat{V}_t > V_t$. Stop macro, if an action leads to a state other than the planned one."}

More formally, let
$$
  S(x) := \arg\max_y (\hat{R}_t(x,y) + \gamma' \hat{V}_t(y) )
$$
be the greedy selection according to $\hat{V}$, and let
$$
  a(x) := \phi(x,S(x)).
$$
Then the macro of state $x$ is the action sequence $a(x), a(S(x)), a(S(S(x))), \ldots,  a(S^n(x)), \ldots$, with the stopping condition "stop at the $n$th step, if either
$\hat{V}_t(S^{n+1}(x)) < {V}_t(S^{n+1}(x))$, or taking action $a(S^n(x))$ in state $S^n(x)$ leads to a state other than $S^{n+1}(x)$.

Naturally, such listing of the macros can be included into the pRL
algorithm, but is not necessary.

\begin{figure}
\frame{\includegraphics[width=12cm]{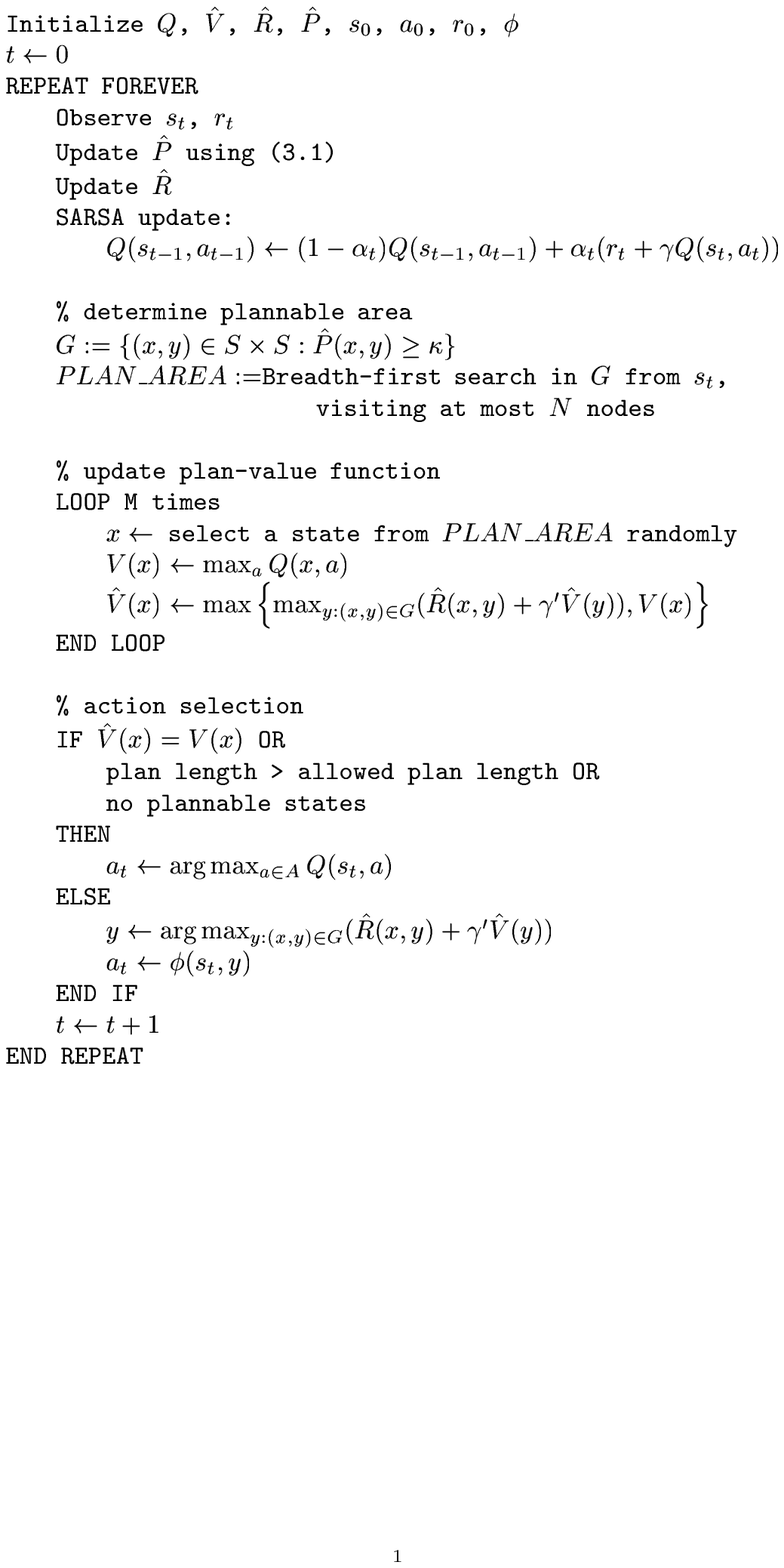}}
 \caption{Pseudocode of the pRL algorithm. In this form, the
algorithm takes at most $\mathcal{O}(N+M)$ steps for every action
selection. \label{fig:pRL}}
\end{figure}

The pseudo-code of the algorithm is summarized in
Fig.~\ref{fig:pRL}.

\section{Near-optimality of pRL Macros} \label{s:optimality}

It is easy to see that the basic value function of pRL converges to the optimal value
function. This follows from the fact that the convergence theorem for SARSA and
Q-learning \cite{singh00convergence} requires (i) sufficient exploration (every $(x,a)$
pair should be visited infinitely often) and (ii) setting the learning rate properly
($\sum_t \alpha_t = \infty$, but $\sum_t \alpha^2_t < \infty$). Clearly, if these
criteria are satisfied with standard RL algorithms, then they are also satisfied in pRL.
However, convergence to the optimal policy is guaranteed only if macros are valued
correctly, i.e., if $\kappa = 1$. The rate of convergence can also be seriously affected
by $\kappa$: $\kappa < 1$ allows for a larger set of macros and
may converge much faster to (possibly suboptimal) solutions. Therefore, it is an
important issue how the learning and the utilization of macros will influence
performance. For the analysis of this problem, an extension of the classical MDP model
is necessary. For this reason, we briefly introduce \emdp s \cite{Szita02Event}, and
review related theorems. The \emdp-theory will be applied to pRL to show that macros are
near-optimal.

\subsection{\emdp s} \label{ss:eps_MDP}

The RL problem can be extended so that the environment is no
longer required to be an MDP, it is only required to remain `near'
to an MDP, i.e. the environment is allowed to change over time,
even in a non-Markovian manner.

The closeness of two environments (which have the same state- and
action-sets) is measured by the distance of their transition
functions. We say that the distance of two transition functions
$P$ and $P'$  is $\epsilon$-small ($\epsilon>0$), if $\| P(x,a,.)
- P'(x,a,.) \|_{L_1} \leq \epsilon$ for all $(x,a)$, i.e. $\sum_y
| P(x,a,y) - P'(x,a,y) | \leq \epsilon$ for all $(x,a)$. (Note
that for a given state $x$ and action $a$, $P(x,a,y)$ is a
probability distribution over $y \in X$.)

A tuple $\langle X, A, \{P_t\}, R \rangle$ is an $\epsilon$-MDP
with $\epsilon > 0$, if there exists an MDP $\langle X, A, P, R
\rangle$ (called the base MDP) such that the difference of the
transition functions $P$ and $P_t$ is $\epsilon$-small for all
$t=1,2,3,\ldots$.

As a simple example for an $\epsilon$-MDP, consider an ordinary
MDP perturbed by a small noise in each time step.

\subsection{Convergence of Learning Algorithms in \emdp s}

One expects that such small perturbations in the environment may not disturb the
performance of the learning algorithms very much. Nevertheless, one cannot expect that
any algorithm finds an optimal value function in an \emdp. (Such a solution may not even
exist because of the perturbations of the environment). However, we shall guarantee that
these algorithms find near-optimal value functions. Formally, the task is to show that
$\lim\sup_{t\to\infty} \| V_t - V^* \| \le \textit{const} \cdot \epsilon$ (or
$\lim\sup_{t\to\infty} \| Q_t - Q^* \| \le \textit{const} \cdot
\epsilon$),\footnote{Unless otherwise noted, $\|.\|$ denotes the max-norm.} where $V^*$
and $Q^*$ are the optimal value functions of the base MDP.

First, consider Q-learning. Recall the update:
\begin{equation} \label{e:eps_Qlearning}
Q_{t+1}(x_t,a_t) = (1-\alpha_t(x_t,a_t))Q_t(x_t,a_t) +
\alpha_t(x_t,a_t)(r_t+\gamma \max_a Q_t(\tilde{y}_t, a)),
\end{equation}
where $\tilde{y}_t$ is selected by sampling, i.e., according to
the probability distribution $P_t(x_t, a_t, .)$.

\begin{thm} \label{thm:eps_Qlearning}
Let $Q^*$ be the optimal value function of the base MDP of the
\emdp, and let $M=\max_{x, a} Q^*(x,a) - \min_{x, a} Q^*(x,a)$. If
\begin{enumerate}
  \item every state-action pair $(x,a)$ is updated infinitely often,
  \item the learning rates satisfy $\sum_{t=0}^{\infty} \chi(x_t\!=\!x, a_t\!=\!a) \alpha_t(x,a) = \infty$ and $\sum_{t=0}^{\infty} \chi(x_t\!=\!x, a_t\!=\!a) \alpha_t(x,a)^2 < \infty$ uniformly w.p.1,\footnote{$\chi$ denotes the indicator function, $\chi(\textit{condition})$ is 1 if the condition is true, and 0 otherwise.}
\end{enumerate}
 then $\limsup_{t\to\infty} \|Q_t - Q^*\| \le \frac{2}{1-\gamma}\gamma M
\epsilon$ w.p.1.
\end{thm}

The proof can be found in \cite{Szita02Epsilon}.

\textbf{The case of \emph{SARSA.}} The proof of the
near-optimality of the SARSA update is similar to the proof of
Q-learning. Here it suffices to refer to
\cite{singh00convergence}: It has been proven for MDPs in
\cite{singh00convergence} that if Q-learning converges to the
optimal value function, then -- under the same conditions -- SARSA
converges, too. The proof in \cite{singh00convergence} carries
over directly to \emdp s \cite{Szita02Epsilon}.

\subsection{pRL in the \emdp\ Framework}

Recall that for finding macro actions, we have applied learning
methods that find optimal policies and value functions \emph{in
the modified environment}, where almost sure transitions (with
probabilities greater than $\kappa$) are treated as sure
(probability 1). It may be worth noting that according to
Eq.~\ref{e:hat_V} the value function of the model (i.e.,
$\hat{V}$) is inherited from the original value function.
$\hat{V}(x)$ can be different from $V(x)$ iff $P(x,y)$ is modified
by the model. Note that modification of $P(x,y)$ is at most
$\epsilon$. Therefore the modified environment is an \emdp\ of the
original one, with $\epsilon = 1-\kappa$, so one can apply the
\emdp-theory to prove the following:

\begin{cor}
The pRL algorithm that uses either DP, Q-learning or SARSA for the
learning of the macro-value function, produces approximations
$\hat{V}_t$ such that $\lim\sup_{t\to\infty} \| \hat{V}_t - V^* \|
\le \textit{const} \cdot (1 - \kappa)$. \label{cor:wpdyna}
\end{cor}
Consequently, the macro actions found by pRL are asymptotically
near-optimal.

\section{Experiments} \label{s:experiments}

\begin{figure}
   \centering
   \subfigure[Immediate rewards of the example labyrinth.
   The colors mark the gained immediate reward when the agent arrives in a the plotted states.
   Reaching the goal state means an immediate +200 reward (this one is not depicted on this picture).]
      {
         \includegraphics[width=6cm]{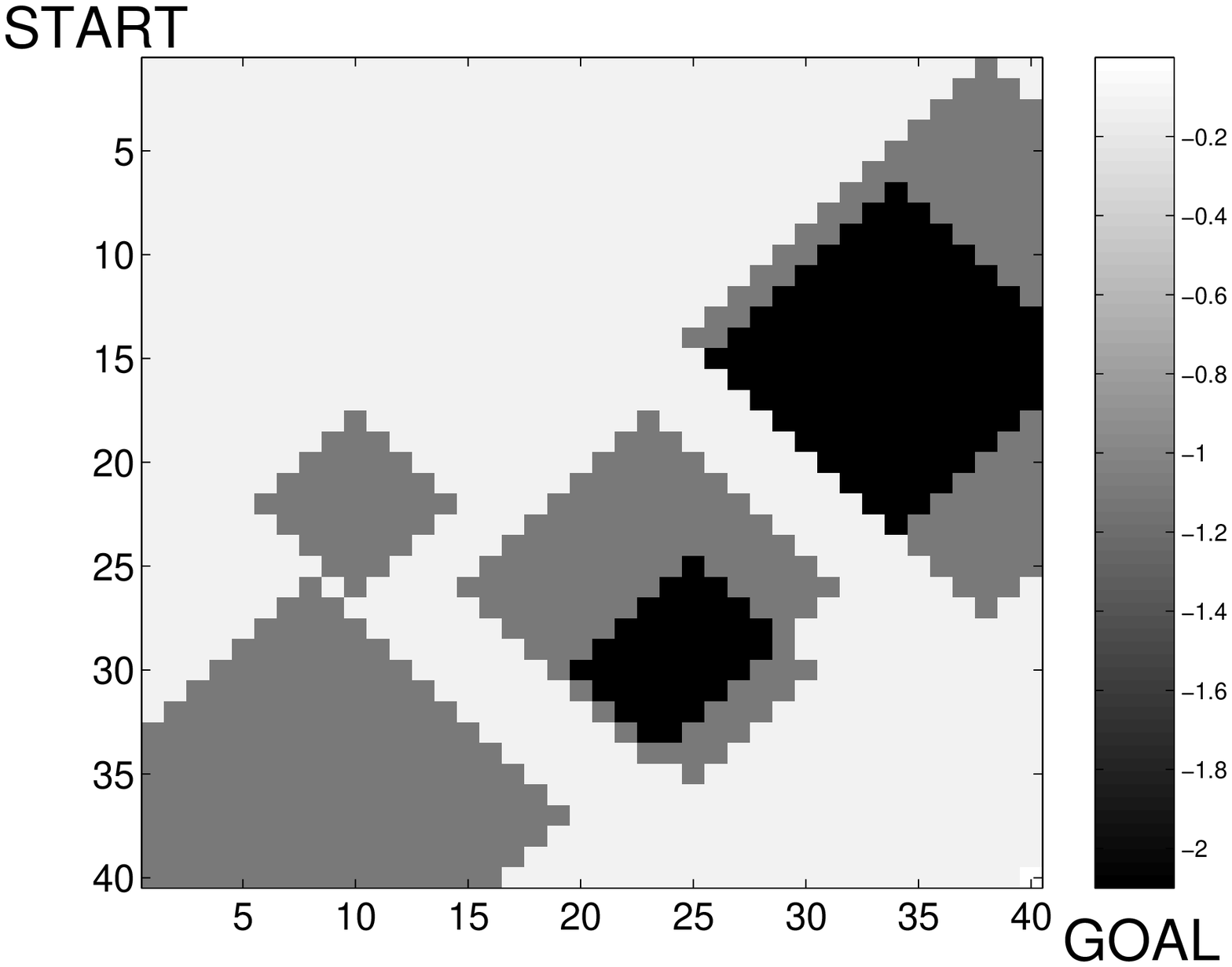}
         \label{fig:maze:rews}
      }
   \subfigure[Probabilities in the example labyrinth.
   The colors mark the probability by which an action leads to the 'good' state from the plotted states, e.g.
   if selecting the action 'NORTH' the agent in a given state, the agent really arrives in the state lying northward.
   If the transition is not successful in this sense, the agent arrives in a 'bad' state chosen with equal probability
   from the other possible next states.]
      {
         \includegraphics[width=6cm]{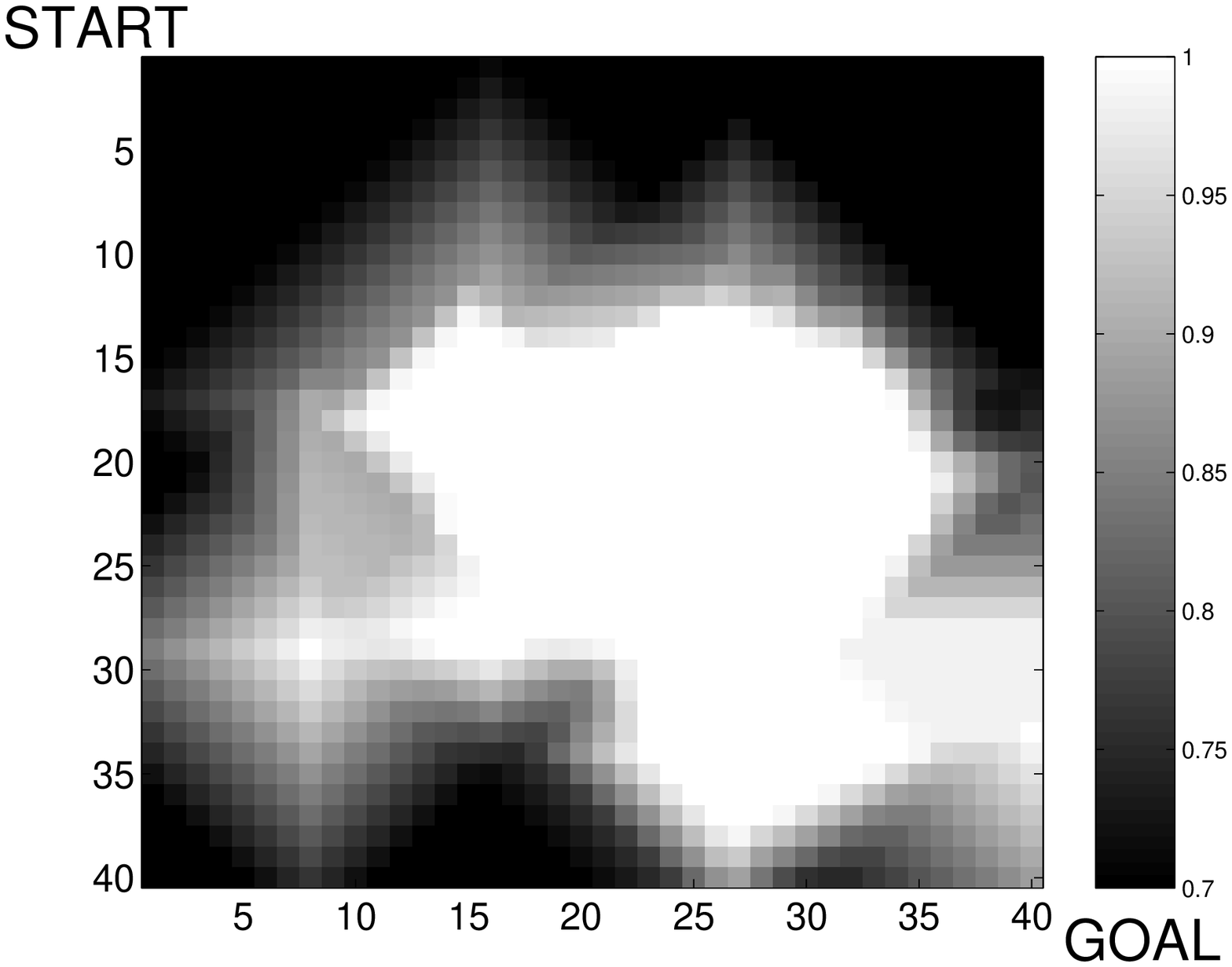}
         \label{fig:maze:probs}
      }
   \hspace{1cm}
   \subfigure[Plannable areas at different $\kappa$]
      {
         \includegraphics[width=6cm]{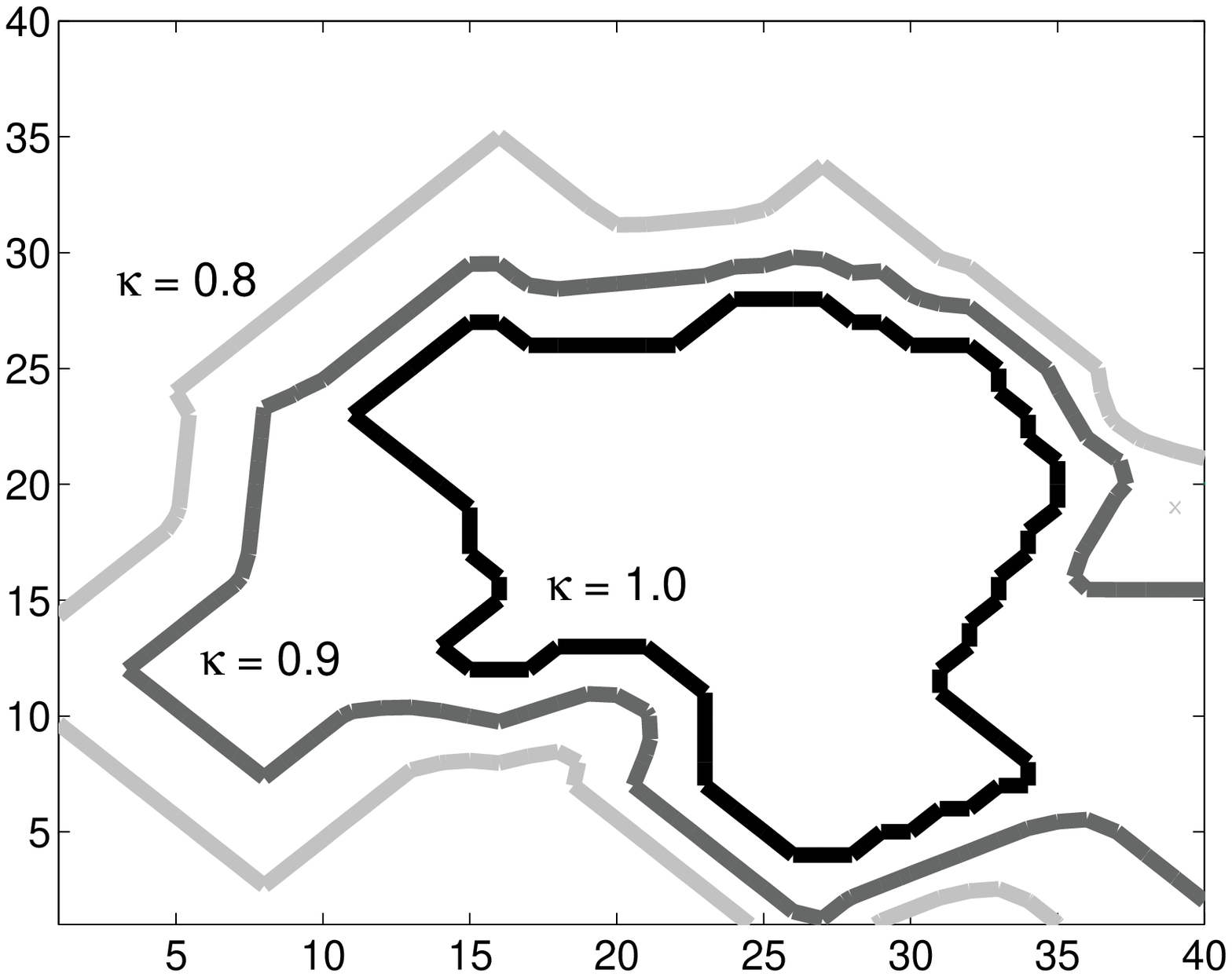}
         \label{fig:maze:kappas}
      }

\caption{\label{fig:maze}}
\end{figure}

\begin{figure}
\centering
\includegraphics[width=10cm]{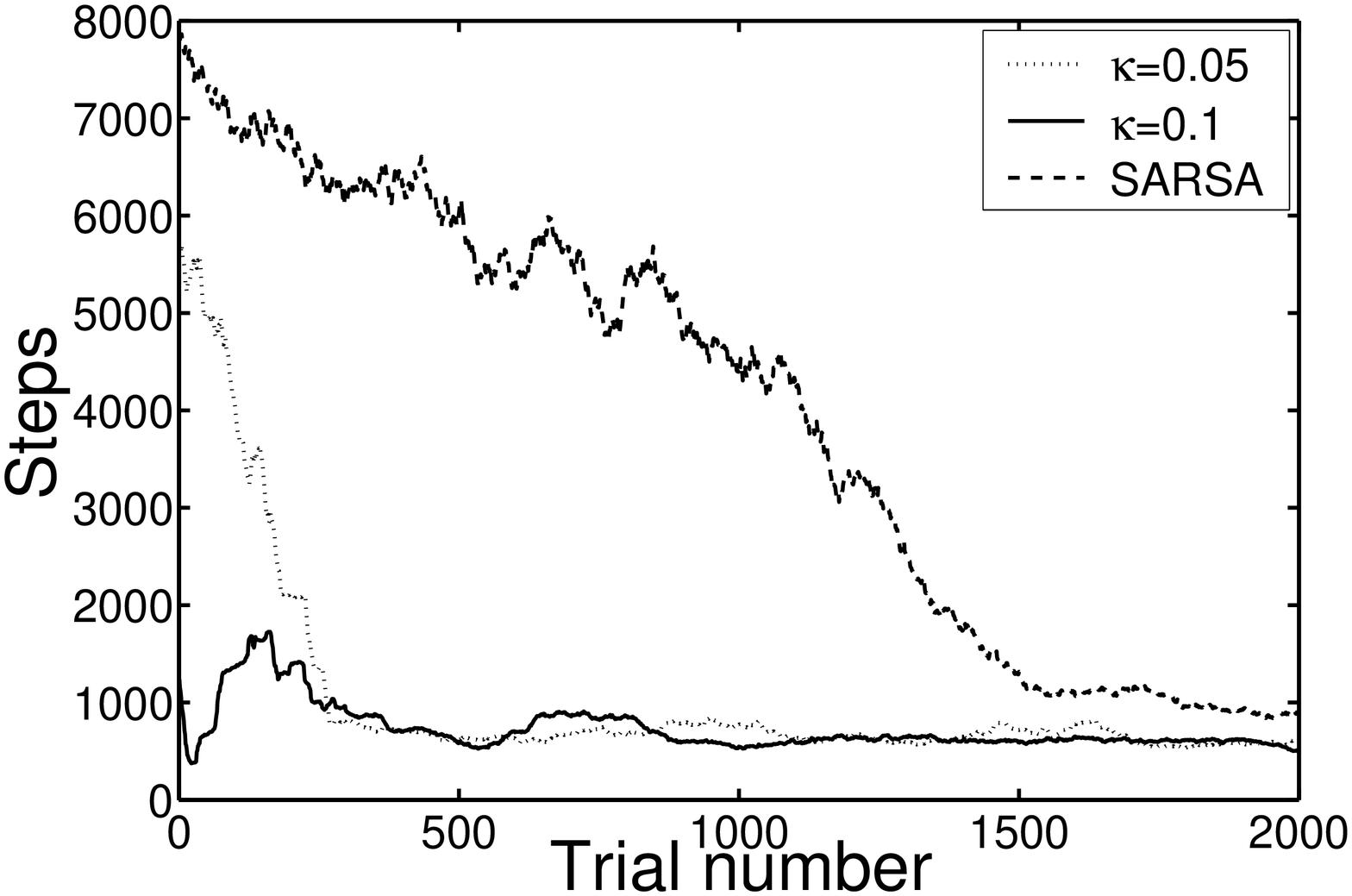}
  \caption{\textbf{Convergence Speed of pRL}
  \newline Learning curves for three different kappa values.
  Dotted line: $\kappa=0.05$, solid bold line: $\kappa=0.15$, dashed bold
  line (SARSA): $\kappa=1.0$.
  The curves represent averaged step numbers using averaging over 500 steps.
  Planning converges much faster then the original SARSA in the
  early phase. Convergence, however, continues beyond
  2000 trials.
  }\label{fig:convspeed}
\end{figure}

\begin{figure}
\centering
\includegraphics[width=10cm]{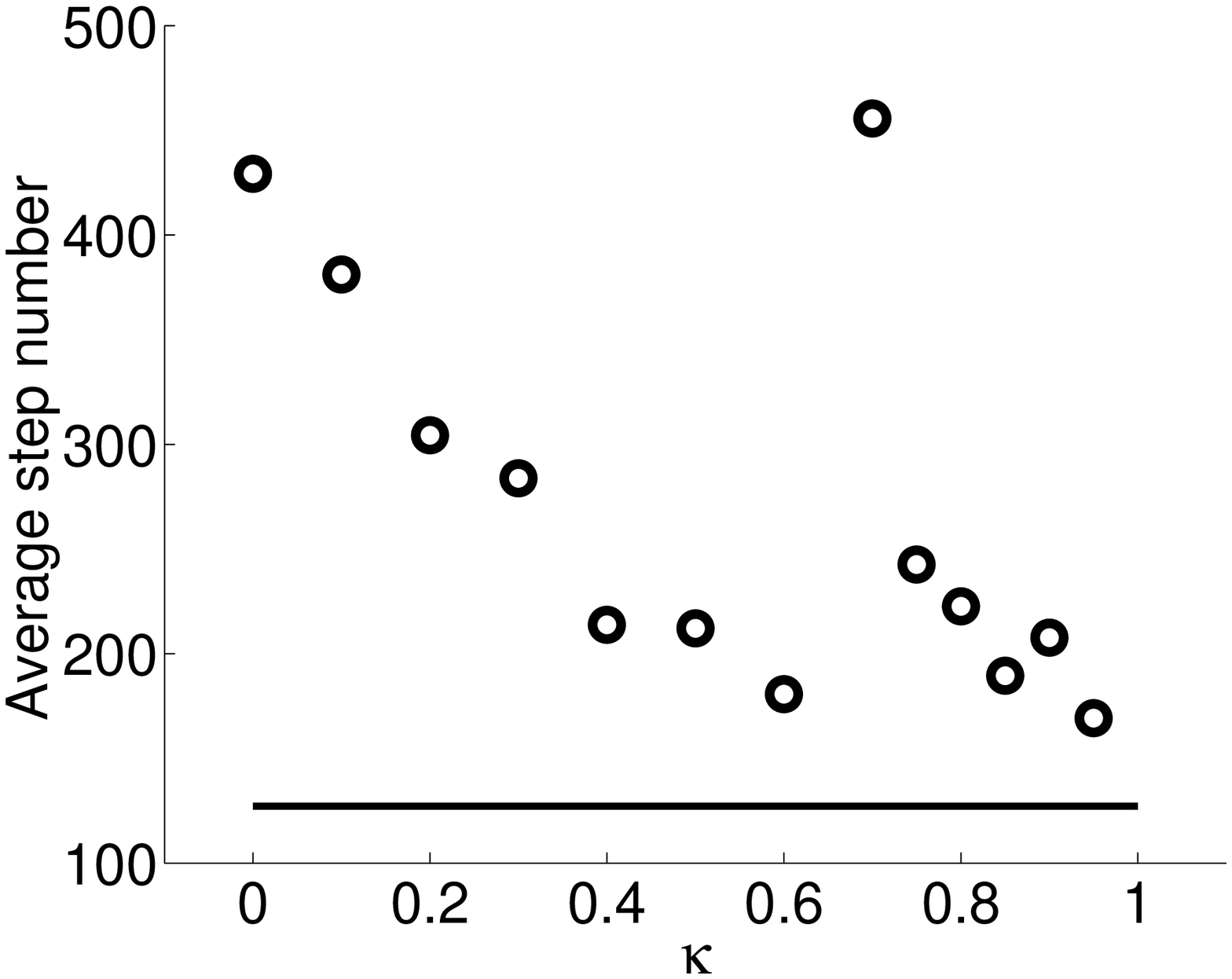}
  \caption{\textbf{Optimality of pRL}
  \newline Final performance of the resulting
  policy as a function of $\kappa$ values. Upon training has converged, the performance of the algorithm
  was measured by averaging the number of steps (in 10000 trials) required to finish one
  trial. The horizontal line depicts the optimal solution found by SARSA
  ($\kappa=1$).
  }\label{fig:performance}
\end{figure}

To test pRL, a $40 \times 40$ maze was generated. The agent starts
in the upper left corner and its goal is in the lower right
corner. In every state, the agent observes its position and can
take four different actions (N(=north),S(=south),E(=east), and
W(=west)). In every state $s$, an action is successful with
probability $P^{succ}(s)$ if the attempted direction is
successfully executed, and it fails with probability
$1-P^{succ}(s)$ when the agent steps to a random, wrong direction.
A lower bound for $P^{succ}(s)$ was set to 0.7. Regions with
higher $P^{succ}(s)$ were generated. The probability of
$P^{succ}(s)$ was gradually increased to 1 within these regions.
The agent received a small $(-0.1)$ negative reward in each state,
except for the goal state, where it received the reward of $+200$.
In addition, some pitfall domains were generated randomly. Every
domain contributed to the local reward by $-1$. In case of
overlaps amongst the domains rewards were cumulated. After
reaching the goal, a new episode was started. An example problem
is shown in Fig. \ref{fig:maze}.

In the experiments, SARSA was utilized with $\epsilon$-greedy
action selection, with eligibility traces
(\cite{Sutton88Learning,singh96reinforcement}). The following
parameters were used: learning rate was held constant at
$\alpha=0.001$. The eligibility decay $\lambda$ was set to $0.95$,
discount factor $\gamma$ was equal to $0.98$, the probability of
random action selection was $0.1$. For pRL, the same setting was
used and several $\kappa$ values were tried. The number of
updates was set to 10.

\subsection{Convergence speed} In theory, by setting $\kappa=1$, pRL
becomes identical to SARSA, and by setting $\kappa=0$, pRL becomes
equivalent to the simplest Dyna algorithm (see \ref{ss:defs}). By
choosing an appropriate $\kappa$, pRL learns at least as quickly
as the better candidate, thus it approximates the convergence rate
of Dyna -- which usually converges in a few trials in this case --
even at low $\kappa$ values (Fig.~\ref{fig:convspeed}). Problems
may arise if it is not known whether the best solution corresponds
to total planning (Dyna) or if no planning is possible at all
(SARSA). By choosing an appropriate $\kappa$, (or a set of
$\kappa$'s) the convergence rate of pRL can always achieve these
boundaries. The method is most useful if computations for
different $\kappa$'s can be afforded.

\subsection{Optimality} The second experiment demonstrates the
near-optimality of pRL. The performance of the resulting policy
for different $\kappa$ values were computed
(Fig.~\ref{fig:performance}). The question is the value of the
constant multiplier of Corollary \ref{cor:wpdyna}; whether it is
too high or not. Here, the assumption about deterministic
transitions does not significantly influence the performance for
high $\kappa$ values, as expected. It was also found that the
resulting policies perform quite well for some fairly low $\kappa$
values; for example, $\kappa \approx 0.5$ performed still
reasonably well. This suggests that better (stronger) estimations
might exist for this problem. However, there is a $\kappa$ region,
which exhibits poor performance. This region is around
$\kappa=0.7$. The poor performance is the consequence of the
special properties of our toy problem. In this $\kappa$ region,
learning is compromised by the fact that most transition
probabilities had a value of 0.7 (except in the plannable
domains), and thus the algorithm had large uncertainties whether a
particular transition is plannable or not.

\section{Discussion} \label{s:discussion}

In this work, we introduced a new algorithm called pRL, which
integrates planning into reinforcement learning in a novel way. An
attractive property of our algorithm is that it can find
near-optimal value function and near-optimal policy under certain
conditions. Furthermore, the macros created by pRL are
near-optimal. Near-optimality is controlled by the certainty of
planning: our algorithm deals with $kappa$-plannable transitions,
which have transition probabilities $0<\kappa \leq 1$.

Our algorithm was illustrated by computer
simulations on a toy problem. It was found that SARSA is slow, but
it leads to optimal solutions. On the other hand, Dyna may be fast
but the resulting policies could be poor. Our algorithm
incorporates both SARSA and Dyna. These are the extremes, SARSA
corresponds to $\kappa =1$, whereas the original form of Dyna
\cite{sutton91planning} is recovered when $\kappa = 0$. By tuning
$\kappa$ in pRL and by using the arising computationally
inexpensive model of pRL, one can quickly find almost optimal
solutions. This feature is most advantageous when off-line
computations are inexpensive. Comparisons amongst the different
methods are provided in Table~\ref{t:comparison}.

\begin{table}
\begin{tabular}{|c|c|c|c|} \hline
   & DP is unexpensive & DP is expensive & DP is expensive \\
   & new data is expensive & new data is unexpensive & new data is expensive
   \\ \hline\hline
  SARSA & slow, optimal & fast, optimal & slow, optimal \\ \hline
  Dyna & fast, optimal & slow, optimal & slow, optimal \\ \hline
  simple Dyna & fast, not optimal & slow, not optimal & slow, not optimal \\ \hline
  pRL & fast, near-optimal & fast, near-optimal & fast, near-optimal \\
      & (low $\kappa$) & (high $\kappa$) &   \\ \hline
\end{tabular}
\medskip
\medskip
\caption{Comparison of RL methods for various problem types.
`Cheap DP' means that DP iterations have low computational cost,
e.g., because the state space is relatively small. `New data is
cheap' means that the agent can easily obtain new experience by
interacting with the environment. `Optimal' means that the method
eventually converges to an optimal policy.}\label{t:comparison}
\end{table}

pRL fits well into existing planning paradigms: it can be seen as
a prioritized sweeping method where only the plannable states
(i.e. states which can be reached almost deterministically from
the current state) are updated. In turn, our method is a
particular (extended) form of prioritized sweeping (see, e.g.
\cite{moore93prioritized}). One can decide to select updates
according to state values \cite{moore93prioritized}, to select
updates ordered by increasing distance from the current state or
by prediction difference \cite{peng93efficient}. Prioritized
sweeping can dramatically improve the performance of planning.
Using our method, one cane also have performance bounds.

pRL can support approaches aiming to solve partially observable
Markov decision processes (POMDPs). Most practical environments
are inevitably POMDPs, which appear to the Markovian agent in the
form of highly non-deterministic transitions. POMDPs are generally
solved by extending the state representation of the agent (e.g. by
estimating the hidden state variables). pRL can be used as a
`pre-filtering' technique. One can decide to use pRL first and to
decouple the almost deterministic domains of the state space,
where -- as a first approximation -- no further extension of the
representation is necessary.

Another promising area for pRL is seen in the novel
\emph{E-learning} method
\cite{Lorincz02Event,Szita02Event,Szita02Epsilon}. E-learning is a
natural candidate for pRL when trajectory planning is desirable.
E-learning is a modification of traditional reinforcement schemes.
In E-learning, the selection of an `action' is equivalent to
selecting a desired next state \cite{Lorincz02Event}. After this
selection is made, the desired state is passed to an underlying
controller. E-learning views the controller (the policy, or an
inverse dynamics, or both) as part of the environment. pRL using
E-learning develops estimations of state-state transition
probabilities. In turn, the set of plannable states is directly
available in E-learning. Thus pRL is greatly simplified within the
framework of E-learning.

In our simulations, the discount rate for the planning ($\gamma'$)
was equal to the discount rate used in SARSA ($\gamma$). A simple
but useful extension arises if one allows for different discount
rates. This could be used for a number of purposes. If $\gamma'
\ge \gamma$ is our choice, then planning may cover larger domains.
On the other hand, $\gamma' \le \gamma$ means that planned actions
will achieve larger short-term rewards. Thus one can tune the
interaction of planning and reinforcement learning by setting the
discount rates, and determine whether to use planning or direct
reinforcement learning to achieve short-term and long-term goals.

\section{Acknowledgments.} This work was supported by the Hungarian
National Science Foundation (Grant OTKA 32487) and by EOARD (Grant
F61775-00-WE065). Any opinions, findings and conclusions or
recommendations expressed in this material are those of the
authors and do not necessarily reflect the views of the European
Office of Aerospace Research and Development, Air Force Office of
Scientific Research, Air Force Research Laboratory.

\bibliographystyle{apalike}
\bibliography{k_plan}
\end{document}